\documentclass{article}
\usepackage{amsmath}
\usepackage{amssymb}

\usepackage{times}
\numberwithin{figure}{section}
\usepackage{graphicx} 
\usepackage{caption}
\usepackage{subcaption}

\usepackage{natbib}

\usepackage{algorithm}
\usepackage{algorithmic}

\usepackage{hyperref}

\usepackage[accepted]{icml2017}

\icmltitlerunning{Submission and Formatting Instructions for ICML 2017}

\begin{document} 

\twocolumn[
\icmltitle{Deep saliency: What is learnt by a deep network about saliency? \\ 
           }




\begin{icmlauthorlist}
\icmlauthor{Sen He}{to}
\icmlauthor{Nicolas Pugeault}{to}
\end{icmlauthorlist}

\icmlaffiliation{to}{Department of Computer Science, College of Engineering, Mathematics and Physical Sciences, University of Exeter,  UK.}

\icmlcorrespondingauthor{Sen He}{sh752@exeter.ac.uk}
\icmlcorrespondingauthor{Nicolas Pugeault}{N.Pugeault@exeter.ac.uk}

\icmlkeywords{Saliency,deep learning,visualization,CNN}

\vskip 0.3in
]



\printAffiliationsAndNotice{}  

\begin{abstract} 
	Deep convolutional neural networks have achieved impressive performance on a broad range of problems, beating prior art on established benchmarks, but it often remains unclear what are the representations learnt by those systems and how they achieve such performance. This article examines the specific problem of saliency detection, where benchmarks are currently dominated by CNN-based approaches, and investigates the properties of the learnt representation by visualizing the artificial neurons' receptive fields. 
	We demonstrate that fine tuning a pre-trained network on the saliency detection task lead to a profound transformation of the network's deeper layers. Moreover we argue that this transformation leads to the emergence of receptive fields conceptually similar to the centre-surround filters hypothesized by early research on visual saliency. 
\end{abstract} 

\section{Introduction}
Deep convolutional neural networks have achieved great success in dealing with computer vision problems, such as image classification \cite{krizhevsky2012imagenet}, object detection \cite{girshick2014rich} and semantic segmentation \cite{girshick2014rich}, etc. However, deep convolutional neural networks are complex non-linear system, and what is learnt by intermediate layers remain in most case mysterious. In addition, despite high performance on many benchmarks, recent published research has demonstrated that despite high performance on benchmark measures, deep networks could be easily fooled by small perturbations of the original signal \cite{moosavi2016universal}, begging the question what are the representations learnt by the networks and how they are used to answer the chosen task (ie, do we recognise a bunny by its ears or by the texture of its fur?). For this reason, it becomes increasingly important for scientists to investigate what is learnt by such networks and what features deep artificial neurons are attuned to, in a way not dissimilar to what neuroscientists did for the human visual cortex. 

\begin{figure}[h]
	\centering
	\begin{subfigure}{0.1\textwidth}
		\centering
		\includegraphics[width=\textwidth]{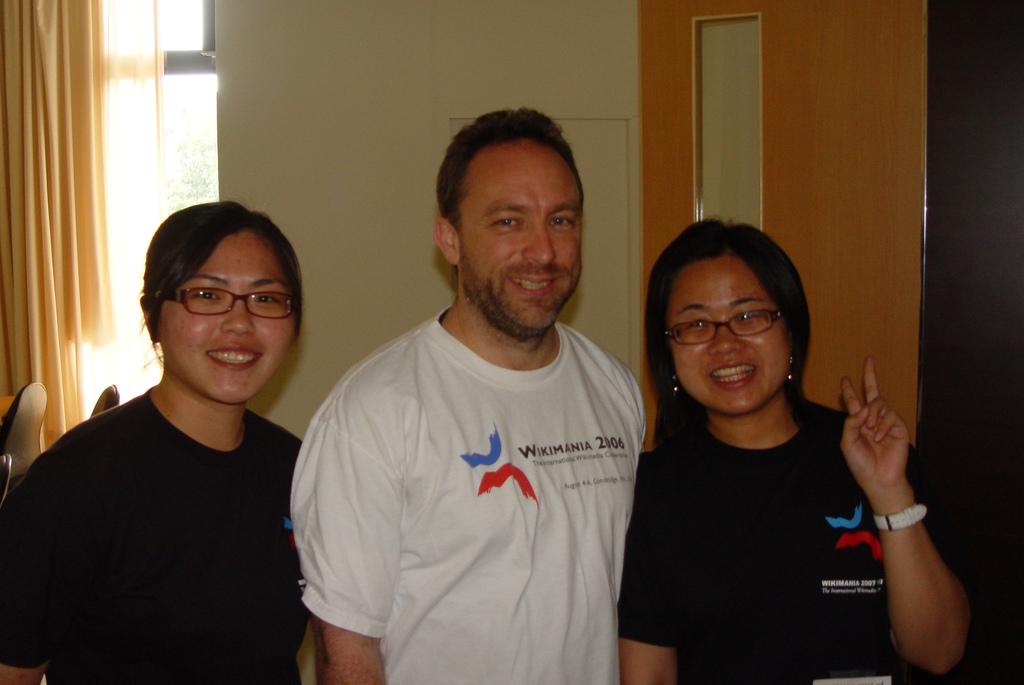}
		\caption{input}
	\end{subfigure}
	\hfill
	\centering
	\begin{subfigure}{0.25\textwidth}
		\centering
		\includegraphics[width=\textwidth]{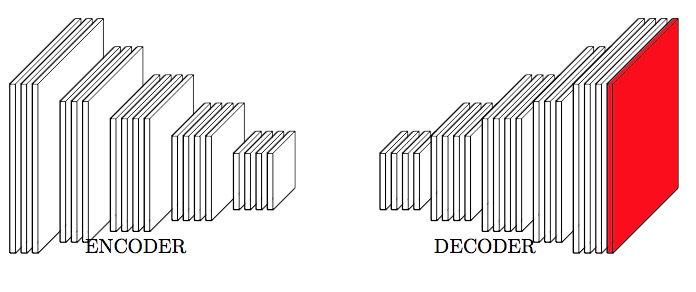}
		\caption{encoder-decoder network}
	\end{subfigure}
	\hfill
	\centering
	\begin{subfigure}{0.1\textwidth}
		\centering
		\includegraphics[width=\textwidth]{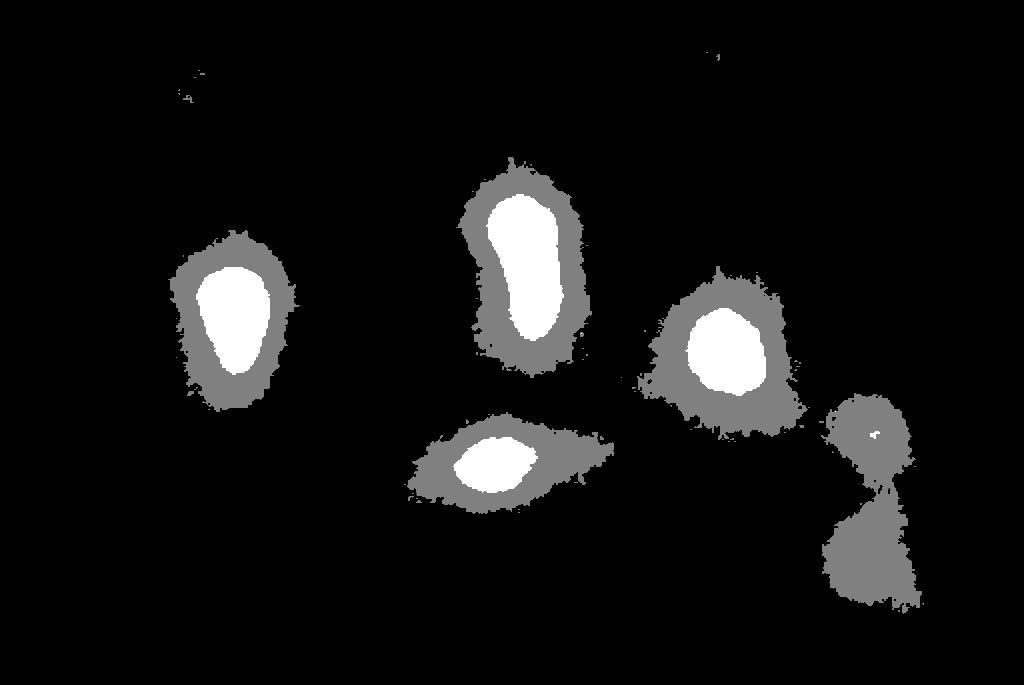}
		\caption{output}
	\end{subfigure}
	\caption{The architecture (the encoder part is fine tuning from VGG-19 convolutional part) we developed for saliency prediction, which is competitive with the state-of-the-art on MIT300 saliency benchmark \cite{bylinskii2015saliency}.}
	\label{fig:modelp}
\end{figure}

In this article, we are concerned with the task of predicting image saliency. Saliency can be defined as how likely a visual pattern is to attract a human viewer's gaze when observing the image. Visual saliency has been the subject of intense study over the last decades, both in psychology and computer vision \cite{borji2013state}, and recent publications have demonstrated that deep neural networks can achieve very high performance on this task  \cite{bylinskii2016should}. We will use a recently developed architecture (see Figure~\ref{fig:modelp}) for saliency detection based on a standard CNN encoding (so-called VGG19 \cite{simonyan2014very}), and visualise the receptive fields of the artificial neurons before and after fine-tuning (the CNN encoder is pre-trained on a standard ImageNet classification task). We demonstrate that after fine-tuning the network for the task, the deeper neurons have evolved vastly different receptive fields to the pre-trained neurons, and display characteristic patterns that evoke the centre-surround difference paradigm hypothesized by early psychophysical research \cite{treisman1980feature}. 

\begin{figure}[h!]
	\centering
	\includegraphics[height=.2\textheight,width=.55\textwidth]{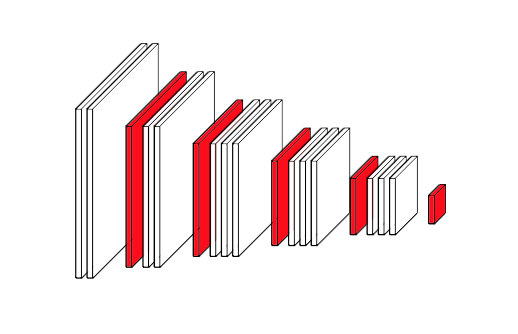}
	\caption{The model used in this paper, we try to visualise the general patterns of the encoder part in Figure~\ref{fig:modelp} that would activate the red pooling layers}
	\label{fig:model}
\end{figure}
\section{Related Work}
With the rise in popularity of deep convolutional neural networks, several groups have recently attempted to visualise which features a network has learned. \citet{zeiler2014visualizing} proposed to back-propagate feature maps obtained by processing a specific image through the network, in order to visualise the image content that activated the feature maps. \citet{yosinski2015understanding} follow a similar approach to develop a tool for deep visualisation, and additionally proposed an approach to visualise features of each layer via regularised optimisation in image space. \citet{nguyen2015deep}, they found that the deep neural networks are easily fooled, and use evolutionary algorithm and gradient ascent method to derive a pattern that the network has a high confidence to determine the derived pattern is belong to a specific class. 

In contrast, we manually clamp the value of single neurons selected from intermediate layers in the network, and back-propagate the activation to the image space, thus deriving the optimal activation pattern for individual selected neurons. 
Hence, this visualisation provides us with an understanding of what patterns the deep representations have become attuned to. 

\section{Methods}
In this work we will be concerned with visualising the input patterns most strongly related to individual neurons in the network. In the following we will call these patterns as the neurons' \textit{receptive fields}, in analogy to biological neurons. 

In deep convolutional neural networks, the forward pass typically consists of three main processes: convolution, non-linearity (usually a ReLU function) and pooling. Similarly, 
the visualisation of neural patterns are produced by three similar processes, in reverse order: that is, upsampling, deconvolution, and non-linearity (again, ReLU). 
We will describe the three processes in turn.

\subsection{upsampling}
\begin{figure}[h!]
\centering
\includegraphics[height=.3\textheight,width=.49\textwidth]{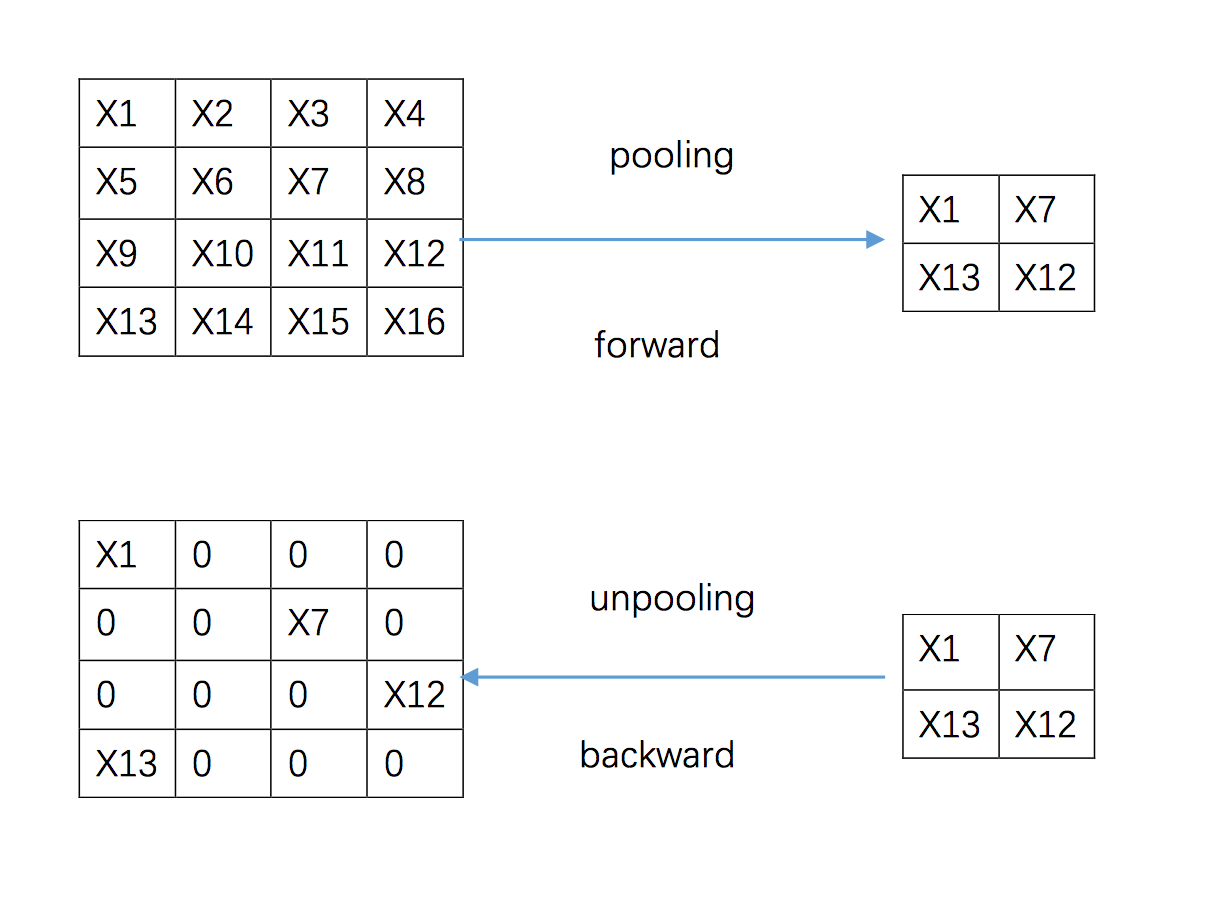}
\caption{classic upsampling method}
\label{fig:classicupsampling}
\end{figure}

The purpose of upsampling is to recover the gradually reduced resolution caused by pooling in the forward pass. The classic upsampling method in feature visualisation is unpooling, using the pooling indices in the forward pass to do unpooling---see Figure~\ref{fig:classicupsampling}. 

\begin{figure}[h!]
\centering
\includegraphics[height=.2\textheight,width=.49\textwidth]{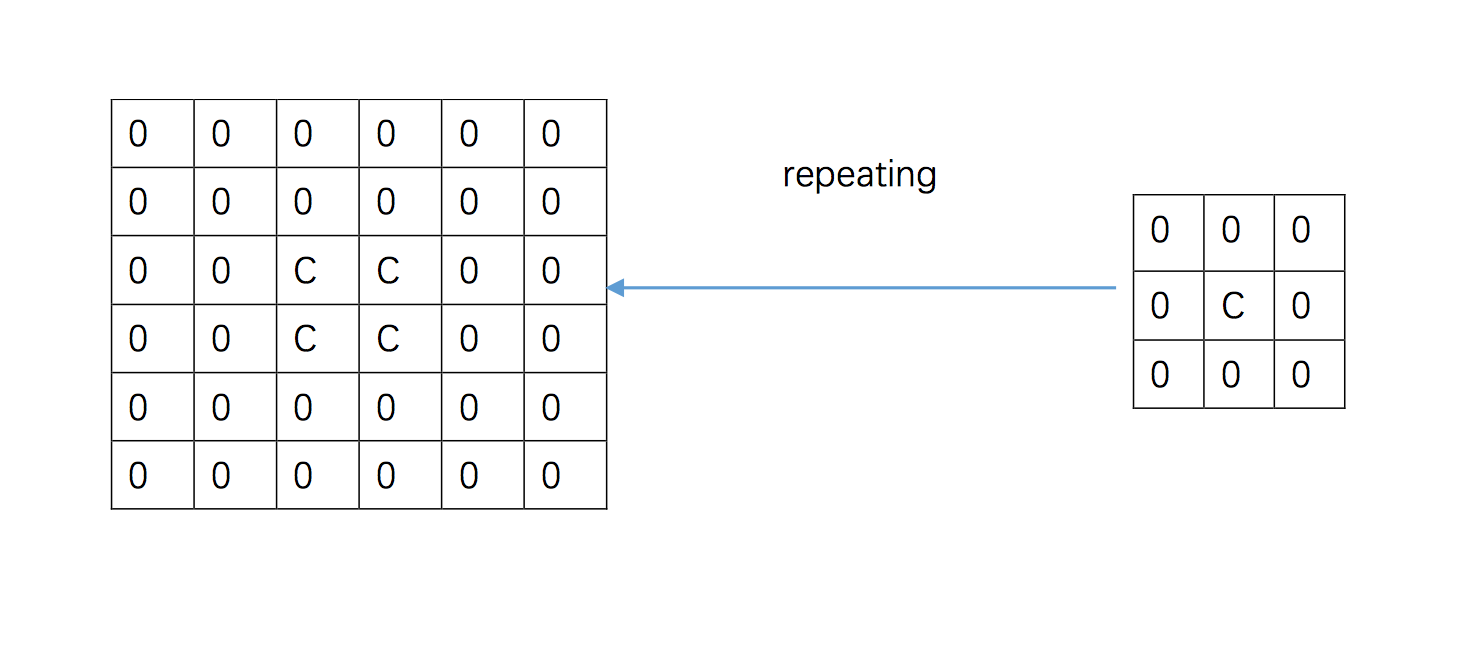}
\caption{upsampling by repeating the sparse feature map, c is a constant}
\label{fig:repeatupsampling}
\end{figure}

Because pooling indices only exist when processing an actual image through the network, these indices are not available when visualizing a neuron's receptive field in abstraction from any input. Hence, in order to visualize individual neuron's receptive fields, we set the pooled feature map as a sparse matrix (with only one non-zero value) and do upsampling by repeating this sparse matrix---see Figure~\ref{fig:repeatupsampling} .

\subsection{Deconvolution}
Convolution is the key process in the forward pass of the convolutional neural network, as it is the one that is tuned by back-propagation during training. It can be formulated as:
\begin{align}
O = I \ast F
\end{align}
where $O$ is the extracted feature map, $I$ is the input and $F$ is the learnt filter. 
Reconstructing the input pattern $I$ which activated an extracted feature map $O$, can be formulated as follows:
\begin{align}
A = O \ast F^T
\end{align}
where $O$ is the extracted feature map in the forward pass, $F^T$ is the transpose of the learned filter and $A$ is the content in the input $I$ which activated $O$.
\subsection{Relu}
the ReLU function in feature visualisation is the same as that in the forward pass of deep convolutional network, which only leave the positive components of the input, and can be formulated as:
\begin{align}
f(x) = \max(0,x)
\end{align}

\section{pattern visualisation}
In this part, we show the general patterns learnt by fine-tuning the network on a saliency prediction task, as well as the patterns for the original VGG-19 network, pre-trained on classification on ImageNet. Additionally, we also visualise individual neurons' receptive field by clamping them and back-projecting to the input domain as described above. 

\subsection{VGG pattern and salient pattern}
\begin{figure}[h!]
	\centering
	\begin{subfigure}{0.23\textwidth}
		\centering
		\includegraphics[width=\textwidth]{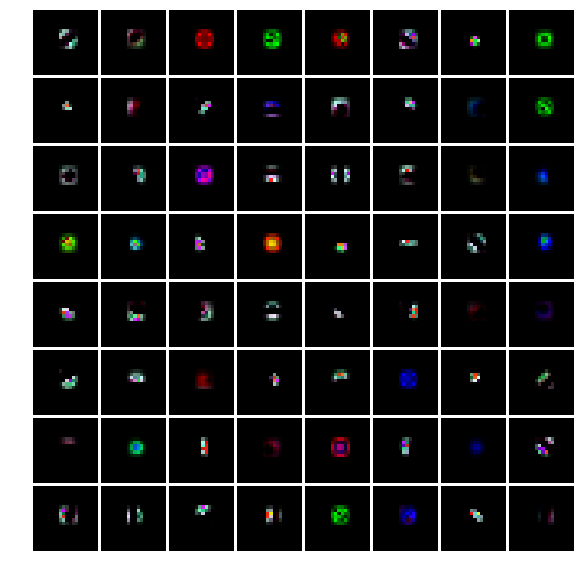}
	\end{subfigure}
	\hfill
	\centering
	\begin{subfigure}{0.23\textwidth}
		\centering
		\includegraphics[width=\textwidth]{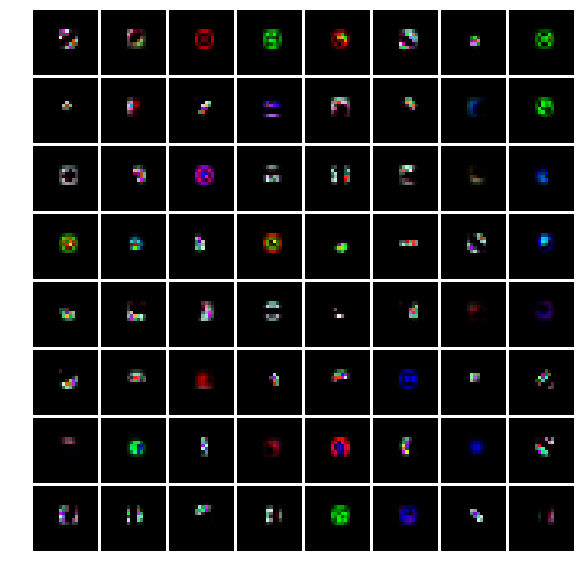}
	\end{subfigure}
	\caption{the 64 general salient(left) and vgg(right) patterns for the first pooling layer}
	\label{fig:general_pattern1}
\end{figure}

\begin{figure}[h!]
	\centering
	\begin{subfigure}{0.23\textwidth}
		\centering
		\includegraphics[width=\textwidth]{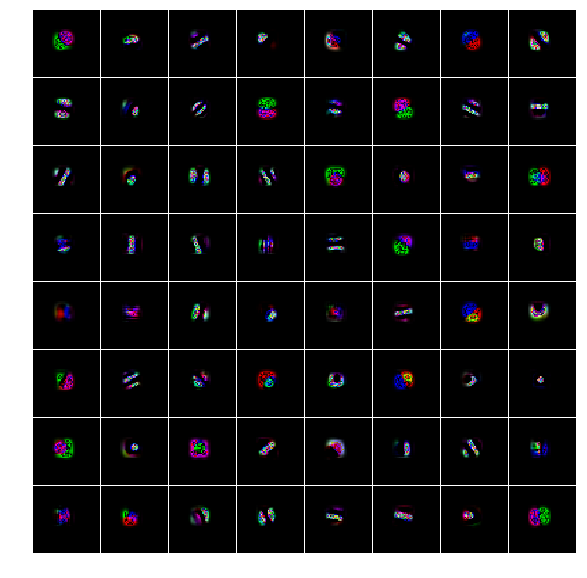}
	\end{subfigure}
	\hfill
	\centering
	\begin{subfigure}{0.23\textwidth}
		\centering
		\includegraphics[width=\textwidth]{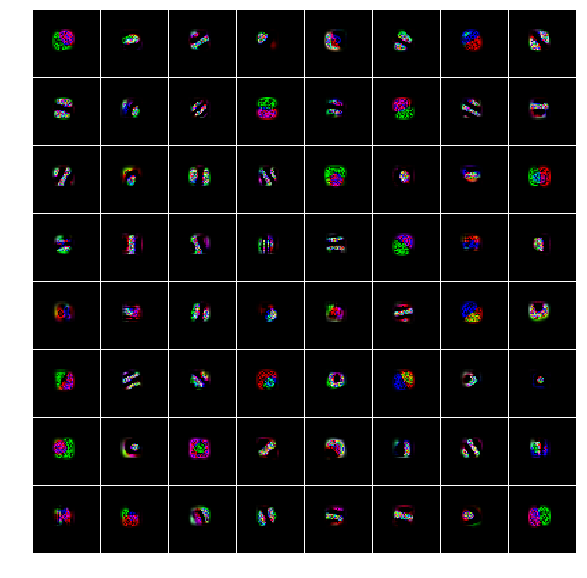}
	\end{subfigure}
	\caption{the first 64 (128 in total)general salient(left) and vgg(right) patterns for the second pooling layer}
	\label{fig:general_pattern2}
\end{figure}

\begin{figure}[h!]
	\centering
	\begin{subfigure}{0.23\textwidth}
		\centering
		\includegraphics[width=\textwidth]{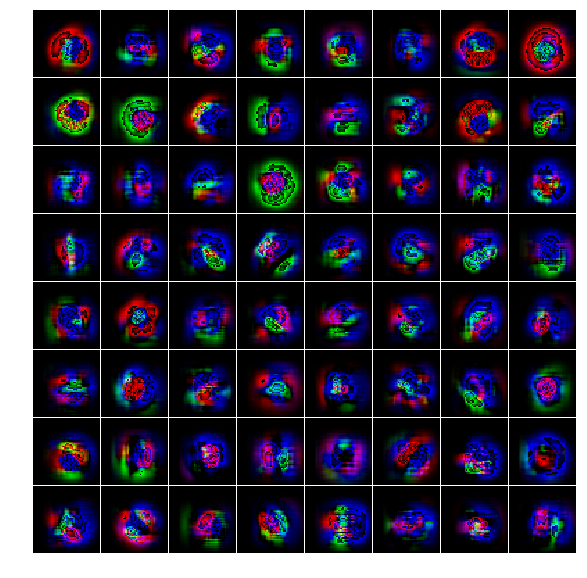}
	\end{subfigure}
	\hfill
	\centering
	\begin{subfigure}{0.23\textwidth}
		\centering
		\includegraphics[width=\textwidth]{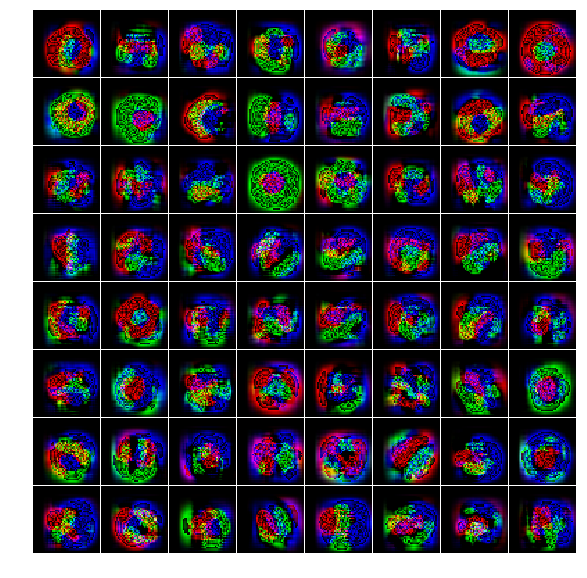}
	\end{subfigure}
	\caption{the first 64 (256 in total) general salient(left) and vgg(right) patterns for the third pooling layer}
	\label{fig:general_pattern3}
\end{figure}

\begin{figure}[h!]
	\centering
	\begin{subfigure}{0.23\textwidth}
		\centering
		\includegraphics[width=\textwidth]{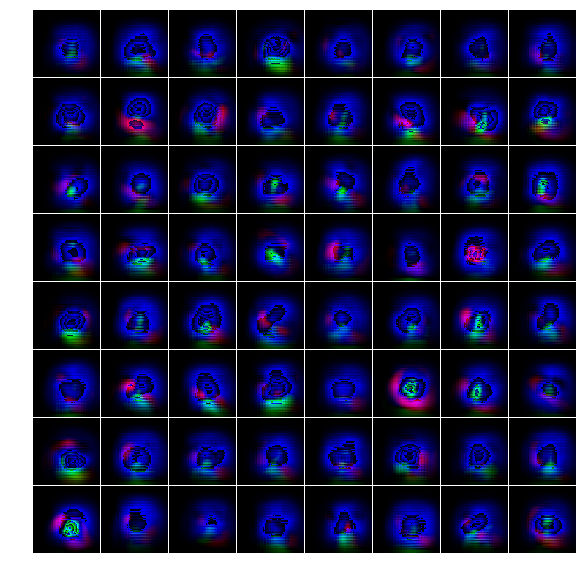}
	\end{subfigure}
	\hfill
	\centering
	\begin{subfigure}{0.23\textwidth}
		\centering
		\includegraphics[width=\textwidth]{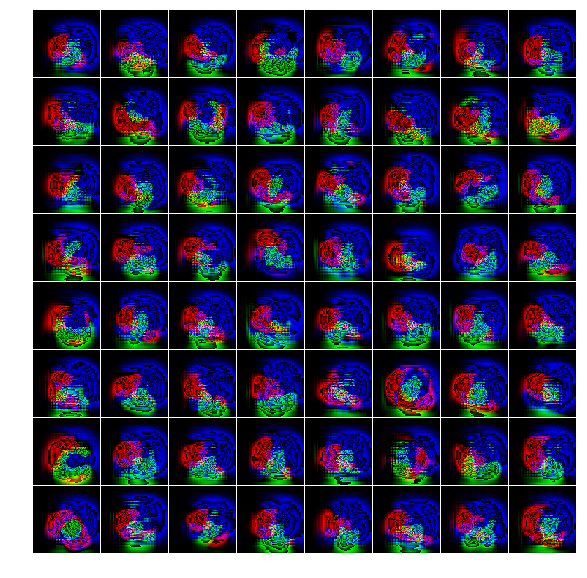}
	\end{subfigure}
	\caption{the first 64 (512 in total)general salient(left) and vgg(right) patterns for the fourth pooling layer}
	\label{fig:general_pattern4}
\end{figure}

\begin{figure}[h!]
	\centering
	\begin{subfigure}{0.23\textwidth}
		\centering
		\includegraphics[width=\textwidth]{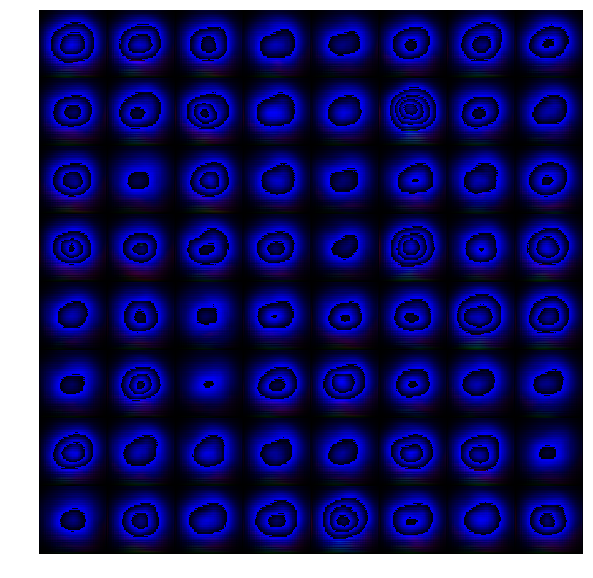}
	\end{subfigure}
	\hfill
	\centering
	\begin{subfigure}{0.23\textwidth}
		\centering
		\includegraphics[width=\textwidth]{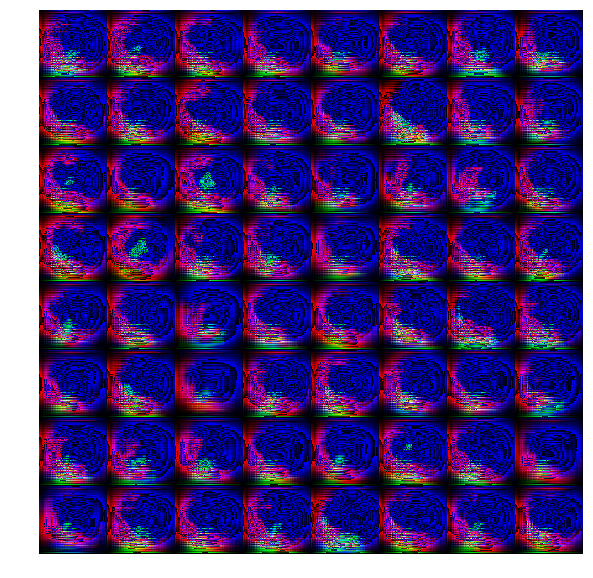}
	\end{subfigure}
	\caption{the first 64 (512 in total)general salient(left) and vgg(right) patterns for the fifth pooling layer}
	\label{fig:general_pattern5}
\end{figure}

Figures~\ref{fig:general_pattern1} to \ref{fig:general_pattern5} contrast the receptive fields learnt by neurons in various layers of the network, both before and after fine-tuning on the saliency detection task.
\\
\\
\\
\\
In those figures, we can see little differences between the neurons' receptive fields in the first three pooling layers after fine-tuning. Some patterns are the same as the edge pattern. However, when considering deeper layers, fundamentally different patterns arise in the neurons' receptive field after fine-tuning for the saliency task: after fine -tuning the deep neurons all appear to have attuned to variations of central-surround patterns. 
Interestingly, such patterns emerge solely through the process of fine-tuning the network, starting from vastly different receptive fields, and they appear to be consistent with theoretical and experimental research on saliency by psychologists.

\subsection{Pattern Propagation}
\begin{figure}[h!]
	\centering
	\begin{subfigure}{0.15\textwidth}
		\centering
		\includegraphics[width=\textwidth]{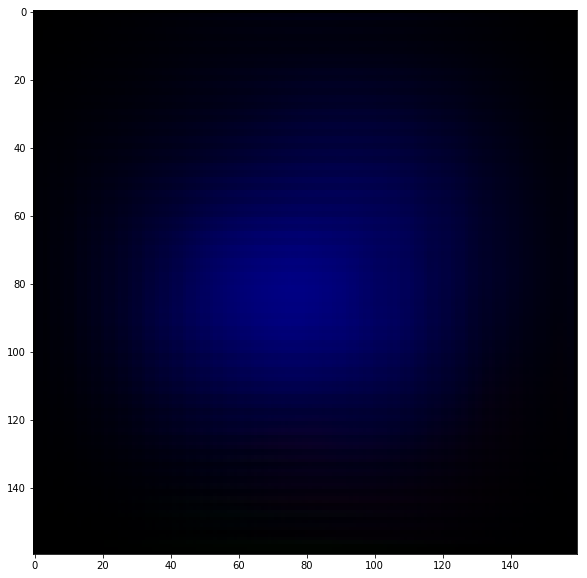}
	\end{subfigure}
	\hfill
	\centering
	\begin{subfigure}{0.15\textwidth}
		\centering
		\includegraphics[width=\textwidth]{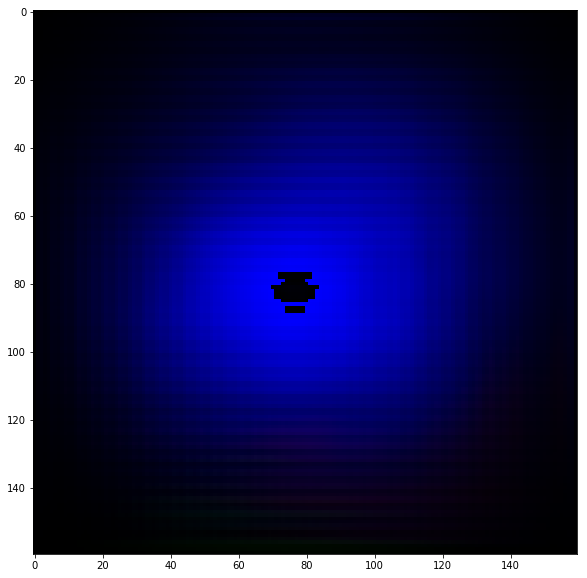}
	\end{subfigure}
	\hfill
	\centering
	\begin{subfigure}{0.15\textwidth}
		\centering
		\includegraphics[width=\textwidth]{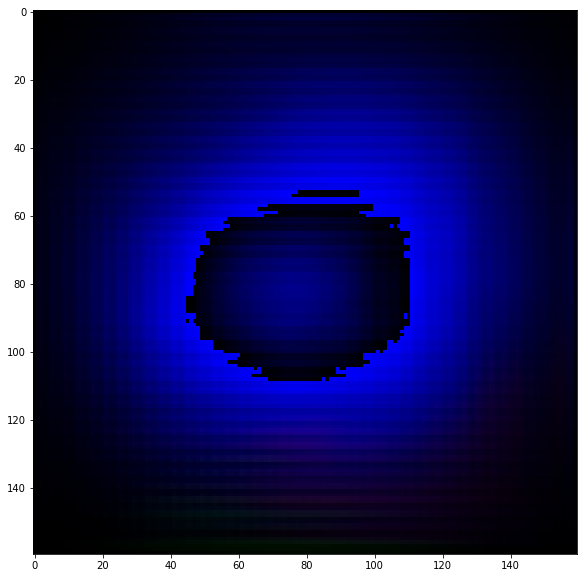}
	\end{subfigure}
	\hfill
	\centering
	\begin{subfigure}{0.15\textwidth}
		\centering
		\includegraphics[width=\textwidth]{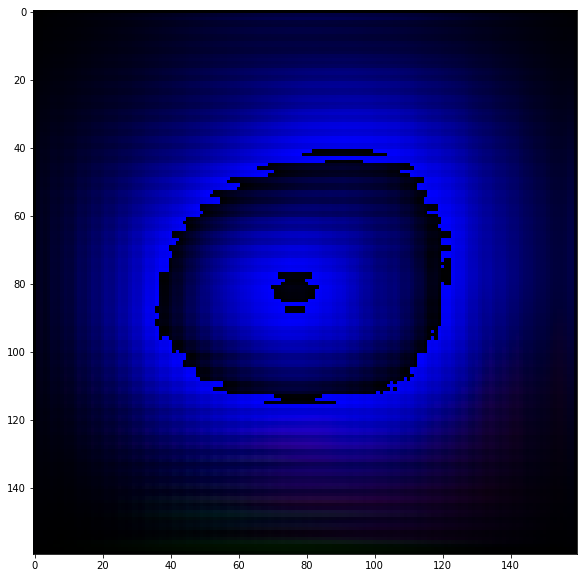}
	\end{subfigure}
	\hfill
	\centering
	\begin{subfigure}{0.15\textwidth}
		\centering
		\includegraphics[width=\textwidth]{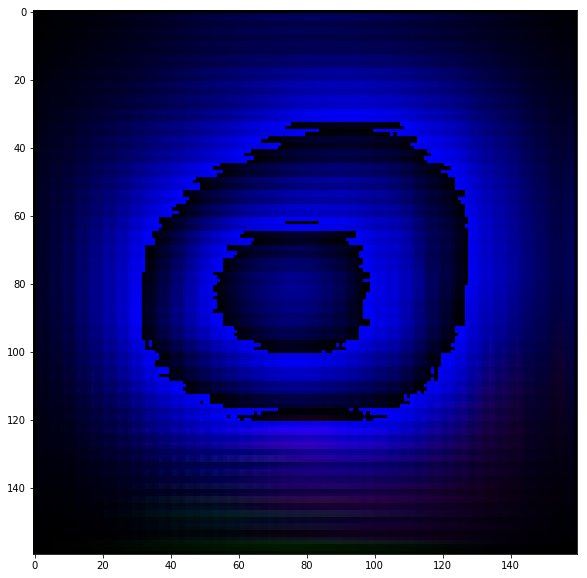}
	\end{subfigure}
	\hfill
	\centering
	\begin{subfigure}{0.15\textwidth}
		\centering
		\includegraphics[width=\textwidth]{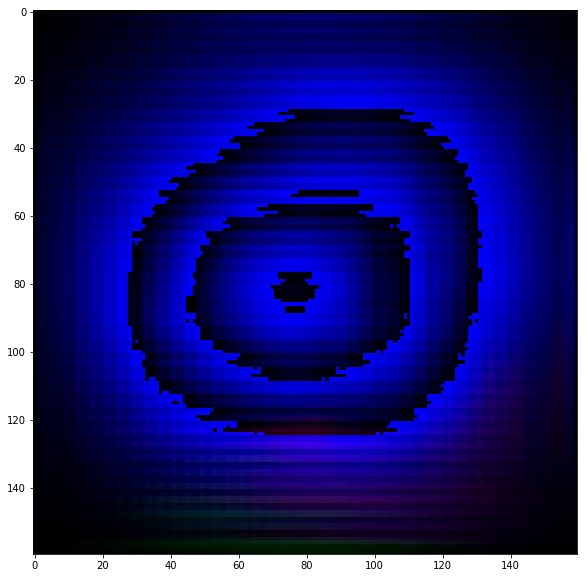}
	\end{subfigure}
	\caption{the pattern propagation by increasing the constant value (from left to right,up to down)}
	\label{fig:propagation}
\end{figure}

In a second experiment, we illustrate how the different patterns yield different levels of activation to specific deep neurons. This is achieved by clamping a neuron-specific output to a constant value from 0.5 to 3. These results are recorded in Figure~\ref{fig:propagation}.
From the figure above, when increasing the neuron's output value its receptive field shows patterns that propagates like the water wave propagation. We argue that the constant in the sparse matrix is the energy of the pattern, the higher energy of a pattern, the wider it will propagate.

\subsection{Pattern Validating}
\begin{figure}[h!]
\centering
\includegraphics[height=.3\textheight,width=.49\textwidth]{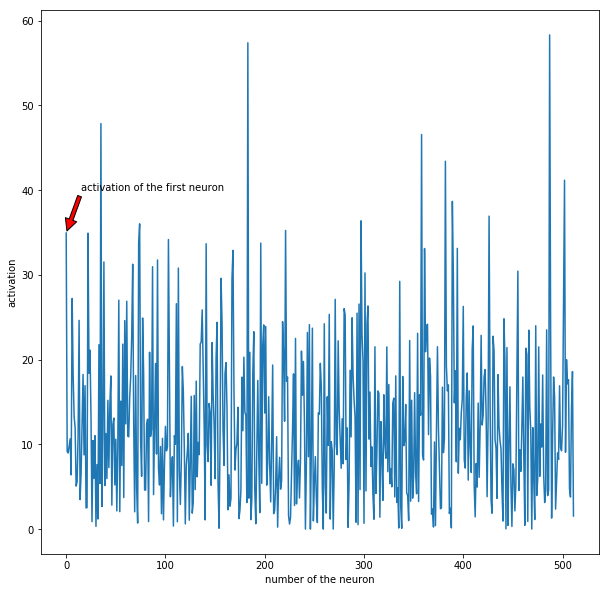}
\caption{the activation of the first general pattern to the fifth pooling layers(512 neurons)}
\label{fig:feature_activation}
\end{figure}

In the previous sections, we have shown that the proposed visualisation strategy can be used to illustrate the patterns learned by deep neurons in a network. In this section we verify that those back-propagated patterns actually activate the selected neuron. An additional question is how such patterns affect other neurons in the same layer. This is tested in a straightforward manner by feeding the back-propagated pattern as an input to the network. Note that due to the pooling process in the forward pass, the resulted pooling feature map may not the same as the clamped sparse matrix used to generate the pattern. We check the activation by the summation of the pooled feature map---see Figure~\ref{fig:feature_activation}.

From this figure, we can see that the first pattern indeed activate the first neuron, as expected. Furthermore, some other neurons have higher activation, demonstrating that the network has developed some redundancy in its coding, whereas other neurons are inhibited. We argue that this is because most of the learned general patterns are similar (central surround difference), some neurons are more sensitive to the central surround difference pattern; and for those neuron that are inhibited is due to the lateral inhibition \cite{ratliff1967enhancement} which is also used in local response normalisation \cite{krizhevsky2012imagenet} when training the deep neural network.

\section{Conclusion}
In this article we proposed a novel approach for visualising the representations learnt by deep neural networks, and specifically to visualise the receptive fields of individual deep neurons. 
We have demonstrated this approach to a VGG-19 network pre-trained on ImageNet classification and fine-tuned for the task of saliency detection. Importantly, we demonstrate that this approach can reveal important insights in what is learnt by the network to achieve its high performance: receptive fields are shown to change drastically from the original VGG-19 representation to characteristic centre-surround patterns. Interestingly, these emergent patterns are consistent with the psychological theories of saliency. This demonstrates that such a visualisation offers an important tool for interpreting the workings of deep neural networks. 

To sum up, by manually set the feature map as a sparse matrix, we derive a set of general patterns for the deep neural network. We also double check the resulted general patterns by forwarding it into the network, which show those general patterns are not illogical and follow the evidence in neurobiology.
\section{Acknowledgements}
This work was supported by the EPSRC project DEVA EP/N035399/1.
\bibliography{example_paper}
\bibliographystyle{icml2017}
\end{document}